\documentclass{article}

\usepackage{arxiv}

\usepackage[utf8]{inputenc} 
\usepackage[T1]{fontenc} 
\usepackage{hyperref}
\usepackage{url}  
\usepackage{booktabs} 
\usepackage{amsfonts}
\usepackage{nicefrac}
\usepackage{microtype}
\usepackage{graphicx}
\usepackage{amsmath}
\usepackage[numbers]{natbib}
\renewcommand{\cite}{\citep}
\newcommand{\textcite}[1]{\citet{#1}}
\usepackage{doi}
\usepackage{tcolorbox}

\title{A Novel Method to Evaluate Models on\\Unreliable, Noisy and Inconsistent Labels:\\Adaptive Resolution Label Aggregation (ARLA)}

\date{} 					% Or removing it

\author{ \href{https://orcid.org/0009-0008-0937-7417}{\includegraphics[scale=0.06]{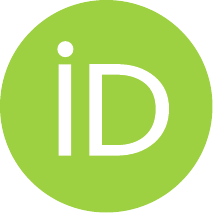}\hspace{1mm}Natasha Randall} \\
	Institute of Information Science\\
	Cologne University of Applied Sciences\\
	Claudiusstr. 1, Cologne, 50678, Germany \\
	\texttt{natasha.randall@th-koeln.de} \\
	\And
	\href{https://orcid.org/0000-0002-1786-8485}{\includegraphics[scale=0.06]{orcid.pdf}\hspace{1mm}Gernot Heisenberg} \\
	Institute of Information Science\\
	Cologne University of Applied Sciences\\
	Claudiusstr. 1, Cologne, 50678, Germany \\
	\texttt{gernot.heisenberg@th-koeln.de} \\
}

\date{}

\renewcommand{\shorttitle}{Adaptive Resolution Label Aggregation (ARLA)}

\begin{document}
\maketitle

\begin{abstract}
Labels are critical for both training and evaluating deep learning segmentation models, but are often inconsistent, noisy, or ambiguous at class boundaries. Many approaches have been developed to support training models on weak labels, but few to none currently exist to facilitate evaluating models on unreliable labels. We therefore introduce a method called `Adaptive Resolution Label Aggregation', or `ARLA', which dynamically adapts the resolution of both the label and the model prediction at inference time before the evaluation metrics are computed. We demonstrate how ARLA can be used to better analyse model behaviour with a practical application to a real flood prediction model, where ARLA was able to overcome issues with inconsistent labelling of forested areas and errors in labels within regions of heavy cloud cover. Our work presents a new approach to evaluating segmentation models, with adjustable parameters to adapt the aggregated resolution to the precision of the label or the level of label noise. Fundamentally, ARLA exploits the information encapsulated by a label but minimises the label error, extracting from the noise a clearer signal of a model's true performance.
\end{abstract}

% keywords can be removed
%%\keywords{First keyword \and Second keyword \and More}

\section{Introduction} \label{intro}
Reliable ground truth labels are fundamental to the development of supervised machine learning or deep learning models, especially for segmentation or image-to-image prediction tasks \cite{fredriksson2020data}. However, labels can be very difficult or expensive to create \cite{rolnick2017deep}, and thus are often generated using (partially) automated processes \cite{ratner2016data}. The gold standard for label creation is from manual human annotation, for example, drawing polygons by hand on remote sensing satellite data to distinguish different land cover classes \cite{hauser2025perfect}. Yet even seemingly gold standard labels frequently have issues; Northcutt et al. \cite{northcutt2021pervasive} estimated an average of at least 3.3\% errors occurring across 10 popular computer vision benchmark datasets that included ImageNet, CIFAR and MNIST, and follow-up work by Wong et al. \cite{wong2022ground} demonstrated that high labeller disagreement led to uncertainty in the reliability and reproducibility of the labels in these datasets.

Although many successful methods have consequently been developed to support \textit{training} models on weak or noisy labels \cite{song2022learning}, there is a significant research gap regarding methodologies for \textit{evaluating} models on unreliable labels. This research gap is a problem, because Northcutt et al.  \cite{northcutt2021pervasive} identified 2916 (6\%) errors in the ImageNet validation subset, and showed how even small numbers of errors in test labels can lead to wrong model selection decisions when based on test accuracy benchmarks. Similarly, Lam and Stork \cite{lam2003evaluating} found that statistically, a classifier with a true error rate of 6\% will report an error rate that is 15\% higher, when only 1\% of the testing labels are incorrect. 

The contribution of our work is a method to evaluate the performance of segmentation models on unreliable, noisy or inconsistent labels, called `Adaptive Resolution Label Aggregation', or `ARLA'. In principle, ARLA works by adapting the resolution of both the label and the model prediction - in accordance with the expected precision of the labels - when calculating evaluation metrics at inference time, without needing to re-train the model. It thus extracts a much clearer signal of both the model's performance and true error, without obscuring either the model's strengths or weaknesses. The following practical examples of implementations of ARLA focus on labels from a flood prediction dataset, as remote sensing is a field in which accurate labels are often very difficult to derive. However, the method can be applied to any segmentation, image-to-mask, or mapping related task.

\section{Related Work}\label{related_work}
The vast majority of previous works relating to unreliable labels have presented only approaches to model training. These include the development of noise adaptation layers in deep learning architectures, regularisation techniques to avoid overfitting to labelling mistakes, and loss function re-weighting based on label or model confidence \cite{song2022learning,karimi2020deep}. These methods have proven relatively successful, in particular because deep learning models are considered to be naturally robust to label noise~\cite{northcutt2021pervasive}; Rolnick et al.  \cite{rolnick2017deep} demonstrated that a standard deep neural network performed well even on training sets in which label accuracy was extremely low. 

In contrast, unreliable and noisy labels have a significant negative effect on model evaluation metrics when used as the ground truth in test datasets \cite{ostmeier2023use}. This impact is well established; Kimhi et al.  \cite{kimhi2024noisy} quantified the influence of different types of annotation errors on segmentation model performance, and Nordström et al.  \cite{nordstrom2022image} analysed the impact of noisy labels on the accuracy and dice metrics for segmentation problems.

The most commonly suggested solution employs methodologies to evaluate models without the use of any ground truth labels at all \cite{lamiroy2011computing, kohlberger2012evaluating}. Other proposed approaches include aggregating labels from multiple different annotator \cite{stutz2023evaluating}, using human-in-the-loop \cite{yang2023assessing}, or applying probabilistic thresholds to estimate noise \cite{northcutt2021confident}. The worst-offending labels can then be cleaned, or simply only the highest quality labels selected to evaluate the models upon \cite{bernhardt2022active}. However these solutions all possess significant disadvantages. Evaluating without ground truth labels at all, or dropping some of the labels, fails to exploit the information that does still exist in even unreliable labels. Cleaning or otherwise improving the labels, whether manually or with automated processes, may simply not be possible when the errors themselves are difficult to quantify, or highly time-consuming for very large datasets.

Cho \cite{cho2024weighted} argues that most model prediction errors occur due to high classification ambiguities around object boundaries, and as such, many works have focused on the borders of segmentation classes \cite{cheng2021boundary, zhang2026uanet}. Ayala et al. \cite{ayala2024guidelines} created a ``tolerance buffer'' at the class boundaries, and then ignored these pixels when calculating the metrics. Soft labels are another popular approach that can be used to blend multiple classes when the class membership of a pixel is ambiguous, as is common at borders \cite{jamali2026softlabels}. Soft versions of metrics, such as the fuzzy F1 score, calculate the model performance directly on the soft labels \cite{harju2023evaluating}. Many of these methods were originally developed under the assumption that the model predictions are inaccurate and that the labels are reliable. Nevertheless, they can still be applied to the problem of unreliable labels, and the two core methodologies - ignoring pixels within a tolerance buffer and evaluating on soft labels at class borders - serve as baseline comparisons with ARLA in section \ref{results}. 

Limited research has been conducted on evaluating classification models on the original unreliable labels. Lam and Stork \cite{lam2003evaluating} utilised a statistical approach to estimate the actual error rate of a classifier based on the probability of the mislabelling rate, and Görnitz et al. \cite{gornitz2014learning} attempted to infer the true label by extracting the structured noise, under the assumption that the label error is not i.i.d.. To the best of our knowledge, no works currently exist on evaluating segmentation models on unreliable labels. 

\section{Problem}\label{problem}
\begin{figure}[!h]
	\includegraphics[width=1\textwidth]{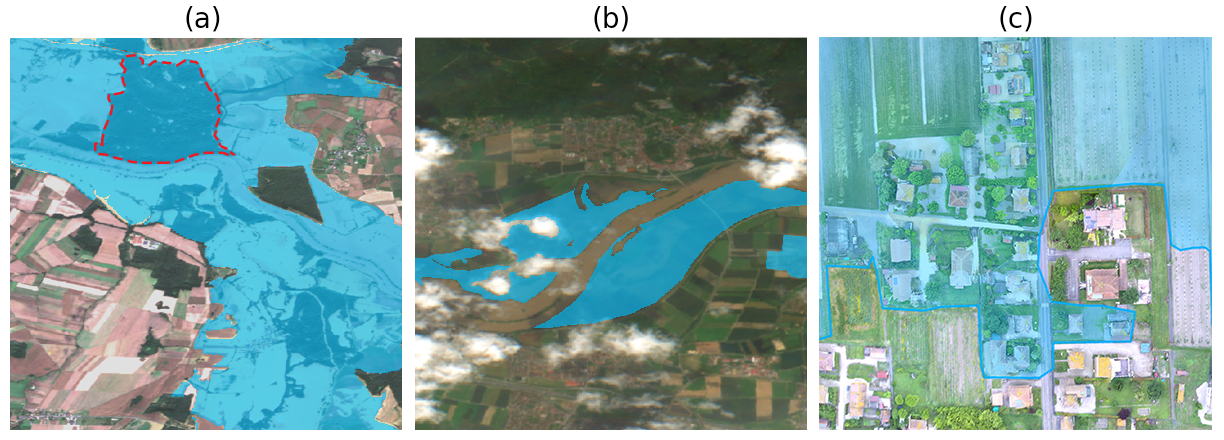}
	\caption{Examples of unreliable, inconsistent, ambiguous and noisy labels in a real flood mapping dataset. The blue overlay corresponds to the binary flood water label.}
	\label{fig1}
\end{figure}

Annotating segmentation class labels from remote sensing data is particularly error-prone, because the annotator is attempting to map a physical property from indirect, imperfect information sources. Three practical examples of common issues are provided in figure \ref{fig1}. These were taken from a real dataset produced by the Copernicus Emergency Management Service~\cite{cems2026}, who provide on-demand mapping of natural disasters such as floods, by annotating the occurrence of flood waters from satellite imagery.

Figure \ref{fig1}a shows an example of inconsistent labelling, where the annotator has inconsistently both included and excluded flooded forests from the flood water segmentation label. The red dashed line highlights one of the areas of forest that was included in the flood label, even though it has similar characteristics to the forest to its bottom-right which was excluded. Whichever of these two behaviours a model trained on this dataset learns, errors will hence be incorrectly reported when the model is evaluated on this label.

In figure \ref{fig1}b, the label is highly unreliable. The annotators avoided labelling areas as flooded where clouds have obscured the underlying reference satellite image, however in actuality flood waters do exist under the cloud cover, expanding out from the western bank of the river. It is often the case that deriving perfect labels is impossible, due to missing satellite data, heavy cloud cover, or poor data resolution. However these discrepancies were not recorded in the metadata, and the reference images were not provided with the dataset.

Figure \ref{fig1}c demonstrates label ambiguity. Even when there are no cloud cover issues, and the reference image is of very high resolution, defining what is the ground truth can be difficult; the boundaries of the flood and non-flood classes are particularly ambiguous. Some noise is also apparent in the label - such as at the middle of the road - and the annotation pipeline's use of polygons has tended to oversimplify complex edges.

Segmentation labels can therefore be unreliable, inconsistent, ambiguous and noisy. Furthermore, users are typically training and evaluating their models on dataset labels created by a third party, and thus lack the resources or the knowledge of the original annotators. Errors in the labels may therefore be completely unknown, and can only be suspected when unusually high model disagreement occurs, where it is difficult to pin down the source. For the examples in figure \ref{fig1}, we were able to obtain the original reference images, but this is uncommon and certainly not guaranteed. In general, the current scientific literature lacks a method to evaluate segmentation models on unreliable labels, even when the label error is unknown, ambiguous or unquantifiable.

\section{Method}\label{method}
In response to the previously described issues, we created the Adaptive Resolution Label Aggregation (ARLA) evaluation method. The goal of ARLA is to extract a clearer signal of both the information encapsulated by the label, and of the model performance. Any mislabelling errors and noise are thus minimised, but the true model error should still be retained.

ARLA is based on a simple but powerful approach. The following implementation describes a segmentation model that is firstly trained at the original label resolution, where the label has a single-channel 2D structure. To evaluate the model performance, the target label $L$, represented as a binary array, is divided into an $s\times s$ grid of contiguous subpatches covering its spatial extent. For each subpatch $S_{a,b}$ the proportion, as a percentage, of the positive class (1) elements from the total number of elements within the subpatch is computed. If this percentage is greater than or equal to the sensitivity threshold $\tau$, then all of the elements in that subpatch are set to 1, else set to 0. The output is the aggregated label $L'$. This process is then repeated for the model prediction $P$, using the same parameter values for the partition and number of subpatches $s\times s$ and the sensitivity $\tau$, outputting $P'$. The evaluation metrics are then applied to $L'$ and $P'$; some examples of this comparison are provided in section \ref{results}. Formally, the aggregation process can be defined as follows:

\newpage
%\noindent\rule[0pt]{\textwidth}{0.5mm}
\begin{tcolorbox}[boxsep=0pt, halign=left, left=5pt, right=5pt]
Let $S_{a,b}$ be a partition covering the complete spatial domain $S$ of the label $L$. Let $f_L$ be a representation of the label $L$ as a binary characteristic function for elements of $S$. Then define
\[
V(S_{a,b}, f_L)=
\begin{cases}
1, & \text{if } \dfrac{\left|\{x \in S_{a,b} \mid f_L(x)=1\}\right|}{\left|S_{a,b}\right|} \ge \frac{\tau}{100},\\[6pt]
0, & \text{otherwise.}
\end{cases}
\]
Thus, $S_{a,b}$ is treated as an element of the binary decision problem over
\[
S'=\{\,S_{a,b}\mid 0 \le a,b < s\,\},
\]
with binary characteristic function $V(S_{a,b}, f_L)$, which gives then $L'$. $V(S_{a,b}, f_P)$ similarly gives $P'$, as they share the same spatial domain $S$. Note that neither the dimensionality of the spatial domain $S'$ nor the specific metric $\left| \cdot \right|$ is fixed, and can be freely varied.
\end{tcolorbox}
%\noindent\rule[4pt]{\textwidth}{0.5mm}

\begin{figure}[!h]
	\includegraphics[width=1\textwidth]{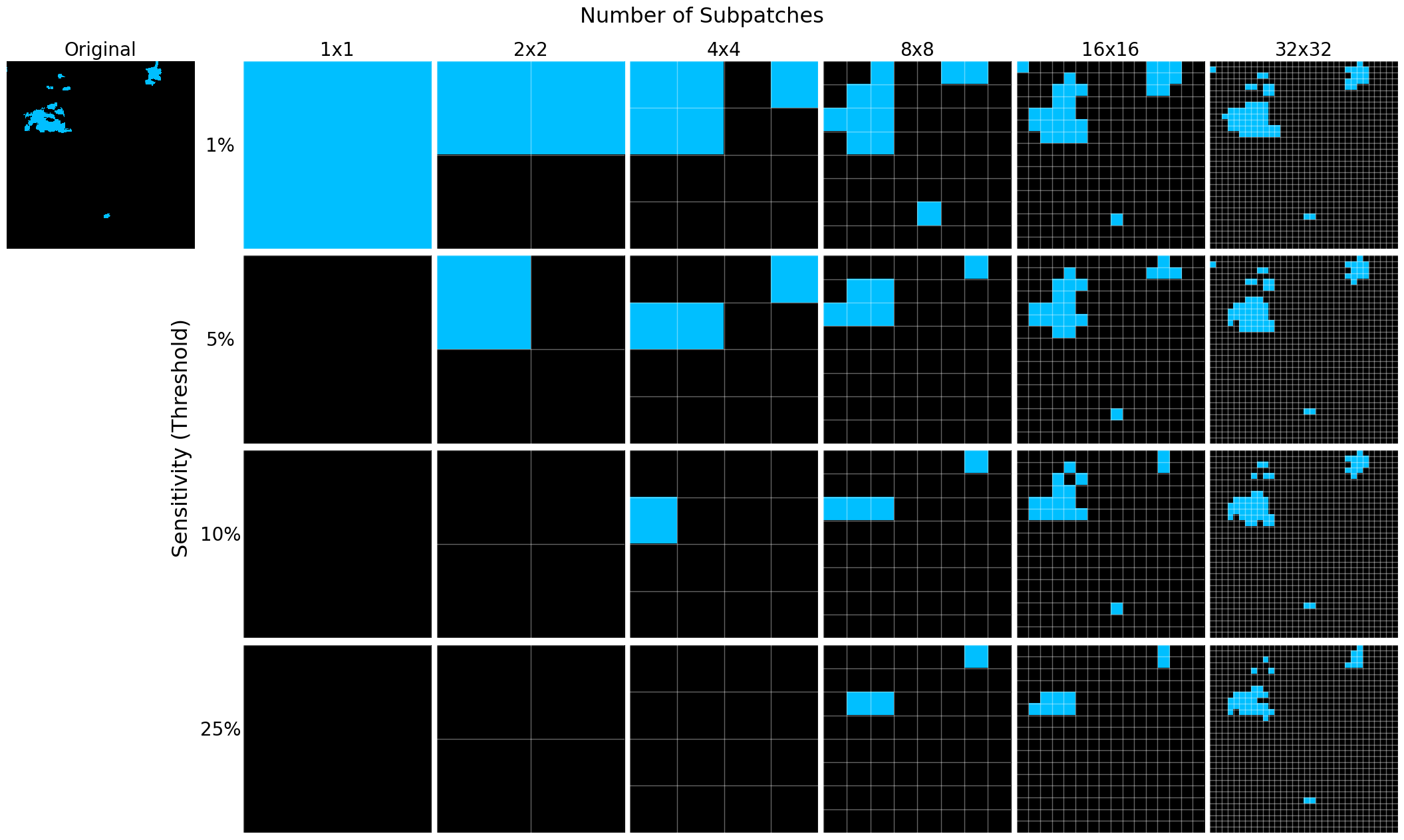}
	\caption{The ARLA method is applied to an example label that depicts areas of flooding (blue: 1) and non-flooding (black: 0), at a variety of different resolutions and sensitivities.}
	\label{fig2}
\end{figure}

For example, if the number of subpatches is set to $1\times 1$, then the segmentation label is essentially converted into a binary classification label. On the other hand, if the original resolution of the label is $256\times 256$ pixels, then equally-spaced $256\times 256$ subpatches would be equivalent to the original segmentation evaluation. As a visual example of the aggregation process on a single label, figure \ref{fig2} demonstrates the effect of different combinations of subpatch numbers and sensitivity parameter values. The original label portrays the segmentation between flood waters (blue) and and non-flood (black), but the drawn class boundaries in this example are very imprecise, and there are a few random blobs - particularly at the bottom of the image - that were mistakenly included during the partially automated generation of the label. However, applying ARLA with parameter values of $8\times 8$ subpatches and a sensitivity of 5\% arguably captures the most relevant information conveyed by the original label; a large patch of flooding occurred at the top left of the image, a smaller patch at the top right, and the erroneous tiny patch of flooding at the bottom of the label is excluded.

\begin{figure}[!h]
	\includegraphics[width=1\textwidth]{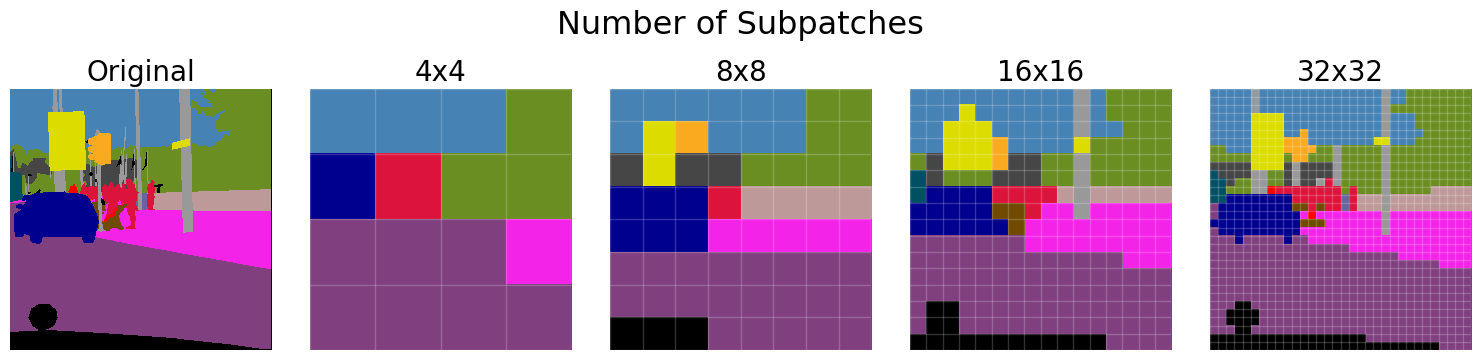}
	\caption{The ARLA method applied to an example multi-class label from the Cityscapes dataset~\cite{Cordts2016Cityscapes}, at four different numbers of subpatches (resolutions).}
	\label{fig3}
\end{figure}

ARLA can also be applied to a multi-class segmentation label, where rather than utilising a sensitivity threshold, the most frequent class is chosen to represent each subpatch. Alternatively, if certain classes are relatively more important than others, then these can be assigned to the subpatch with priority if defined class threshold levels are reached. In figure \ref{fig3}, of an example label from the Cityscapes dataset \cite{Cordts2016Cityscapes}, even at only $8\times 8$ subpatches, the aggregated label already well encapsulates the core information and spatial distribution of the original classes, while filtering out the noise.

\section{Results}\label{results}
\subsection{Example Applications} \label{example_applications}

\begin{figure}[!h]
	\includegraphics[width=1\textwidth]{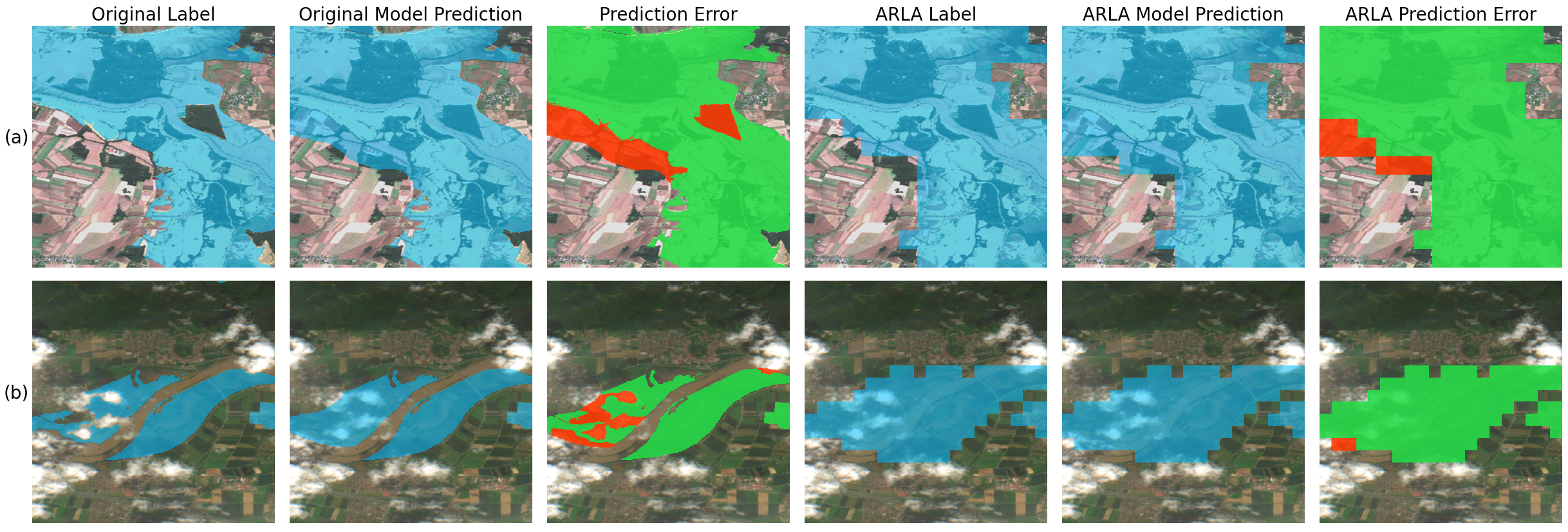}
	\caption{Two examples that compare a label, model prediction, and prediction error, both before and after ARLA has been applied. Blue represents the flood water class, red where the prediction and label differ, and green where they match.}
	\label{fig4}
\end{figure}

Figure \ref{fig4} demonstrates how some of the labelling issues first introduced in figure \ref{fig1} can be mitigated by using ARLA. The original label and a synthetic exemplar model prediction of flood waters (in blue) are compared, where the prediction error is highlighted in red for differences, and green for correct matches. The comparison is then repeated after ARLA has been applied to both the label and the model prediction, with a sensitivity of 1\% and $13\times 13$ subpatches or $20\times 20$ subpatches for figures \ref{fig4}a and \ref{fig4}b respectively.

After ARLA has been applied to figure \ref{fig4}a, the forested region that was inconsistently labelled by the annotator is now included in the positive flood waters class, and the prediction error has consequently improved. Yet the reduction of the resolution has also reduced the precision of the class boundary delineations, seemingly inducing additional errors in the ARLA label. However, in the locations where the model prediction followed the boundary borders in the original label perfectly, both label and prediction will be converted to the exact same new representation after ARLA has been applied - because both are aggregated using the same parameters - and thus no additional error is created. ARLA produces a less noisy, more informative performance signal, but does not inherently obscure model errors; where the model over-predicted flooding in the centre-left of figure \ref{fig4}a, this error is still preserved after applying ARLA.

In figure \ref{fig4}b, the annotator incorrectly excluded flooded regions from the label that were obscured by cloud cover in the original reference satellite image. However, a trained model that had learned to outperform this label would predict these regions as flooded, which would subsequently be recorded as a model error when evaluating on the original segmentation. Yet after ARLA is applied to figure \ref{fig4}b, the model is better recognised to have made a generally correct prediction.

\subsection{Baseline Comparison}

The two methods that were first introduced in section \ref{related_work} are used as baselines for comparisons with ARLA. The first method creates a wide buffer that is centred on the class boundaries, and then ignores these pixels entirely when calculating the metrics. The second method smooths the hard labels to generate soft labels, where the values linearly decay from [1.0, 0.0] (or [0.0, 1.0]) to [0.5, 0.5] as the distance to the class border decreases. Both of the baseline methods focus on the boundaries of classes, as this is one of the few information signals from a label that is guaranteed to exist even without further knowledge from the annotators. In the example applications from figure \ref{fig4}, the forest and cloud cover label errors inherently automatically occur at the borders of the flood and non-flood classes, thus the baseline methods represent reasonable approaches.

\begin{figure}[!h]
	\includegraphics[width=1\textwidth]{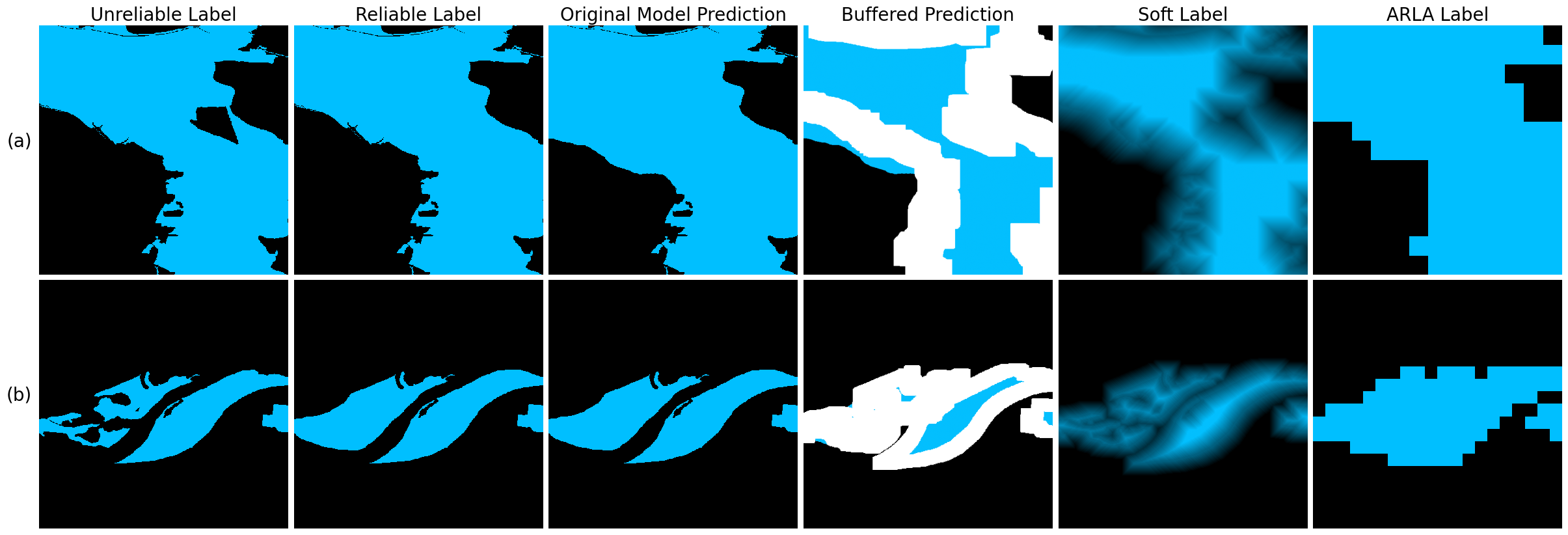}
	\caption{The `Buffer', `Soft Label', and ARLA methods are applied to two examples of unreliable labels, where blue and black represent the two classes, and white the ignored pixels. The corresponding reliable (fixed) label and original model prediction are also shown.}
	\label{fig5}
\end{figure}

Figure \ref{fig5} visually demonstrates the application of the methods to the two examples. The parameters for all of the methods are calibrated to the size of the label error (here, the size of the missing forest and cloud cover areas). The white band in the `Buffered Prediction' designates the pixels that are excluded from the metrics calculation, whereas the `Soft Label' shows the fuzzy values for the flood class. Table \ref{baseline_table} presents the precision of the model as evaluated on a fixed `reliable' version of the label (as can be seen in figure \ref{fig5}). The precision is then calculated after each of the three methods have been applied to the unreliable label; the corresponding difference to the reliable label metric is displayed in brackets. The methods should not simply maximise the model performance, but rather report its true performance, including the true model error, as represented by the model precision on the reliable label with no label errors. Figure \ref{fig5}b is an example of the case where the model matches the reliable label perfectly, but in figure \ref{fig5}a the model makes an over-prediction, and so its reported precision should reflect this.

\begin{table}[!h]
    \centering
    \begin{tabular}{cc|ccc}
         Example & Reliable Label & Buffer & Soft Label & ARLA \\
         \hline
         (a) Forest & 0.908 & $0.945\text{ } (+0.037)$ & $0.760\text{ } (-0.148)$ & \textbf{0.938 ($+$0.030)} \\
         (b) Cloud & 1.000 & $0.979\text{ } (-0.021)$ & $0.618\text{ } (-0.382)$  & \textbf{0.981 ($-$0.019)}\\
    \end{tabular}
    \caption{The model precision as calculated on the reliable label, and after applying the three methods to the unreliable label. The value in brackets is the corresponding difference to the reliable label precision.}
    \label{baseline_table}
\end{table}

After calibrating the buffer method to the size of the unreliable label error, a huge proportion of the label is lost, and the metrics are calculated using only a very small area of the remaining label. In addition, the region with the model error in figure \ref{fig5}a is almost entirely contained by the buffer and hence ignored. Therefore although the precision of the buffer method in table \ref{baseline_table} is very close to the reliable label precision, figure \ref{fig5} reveals that these results are highly misleading. Although the soft labels are a more sophisticated approach, this method can only handle small class border discrepancies well. Not only are the label error regions still not fully excluded, but the class boundaries where the label is actually accurate also lose a lot of precision, resulting in very poor metrics in table \ref{baseline_table}.

ARLA performs better than the two baseline methods on both of the examples. ARLA is capable of handling large label errors without adversely affecting the areas of the label that do not have errors, whereas the baseline methods struggled with both of these aspects. ARLA can also be adapted to smaller, more fine-grained class border discrepancies by simply increasing the number of subpatches, as shown in figure \ref{fig2}.

\subsection{Flood Event Application}

The following application example relates to a severe flood event that occurred in May of 2023 in Emilia-Romagna, Italy, which was mapped by the Copernicus Emergency Management Service, as depicted by the blue label in figure \ref{fig6}a. A flood prediction model was trained on this event to learn to forecast the locations of flooding. The model was input with the antecedent conditions and general information describing the region - such as precipitation and topographical features - and outputs a flood water segmentation. Figure \ref{fig6}b shows the differences between the label and the model prediction, where green indicates a correct match, red an over-prediction by the model, and yellow an under-prediction. The black boxes depict the extent of the $256 \times 256$ pixel patches that were individually input into the model, with the outputs then spatially mosaicked together to form the full flood event. The satellite image underneath is the reference that was used by the annotators to create the original label.

Figure \ref{fig6} is an extreme example of highly unreliable labels. Some of the segmentation labels are wrongly missing due to the heavy cloud cover, the class boundaries are inconsistent and imprecise, and there are noisy labels that were generated using an alternative automated process, due to missing satellite reference imagery. However, figure \ref{fig6}b demonstrates how deep learning models are considerably robust to handling unreliable labels during training; by extracting the patterns from the noise, the model has still learned to (correctly) predict that flooding should occur underneath the clouded areas. Yet as a result of the label errors, during evaluation, the model is calculated to have a low precision of 0.52, evidenced by all of the red (ostensibly) over-prediction in figure \ref{fig6}b.

In figure \ref{fig6}c, ARLA has been applied to each of the label and model prediction patches, using a particularly coarse partitioning resolution with parameters of $1 \times 1$ subpatches and sensitivity of 2.5\%. In the upper-left of the flood event, ARLA has significantly minimised the amount of cloud-induced red over-prediction of the original segmentation, instead recognising the model's robust performance with much more green. When metrics are calculated using the ARLA versions of the label and prediction, the model precision subsequently increases from 0.52 to 0.80. ARLA has also not simply obscured the weaknesses of the flood prediction model. The model clearly under-predicts flooding in the bottom-right of the flood event, thus the model recall is still only 0.63 after ARLA has been applied. 

\begin{figure}[!h]
	\includegraphics[width=1\textwidth]{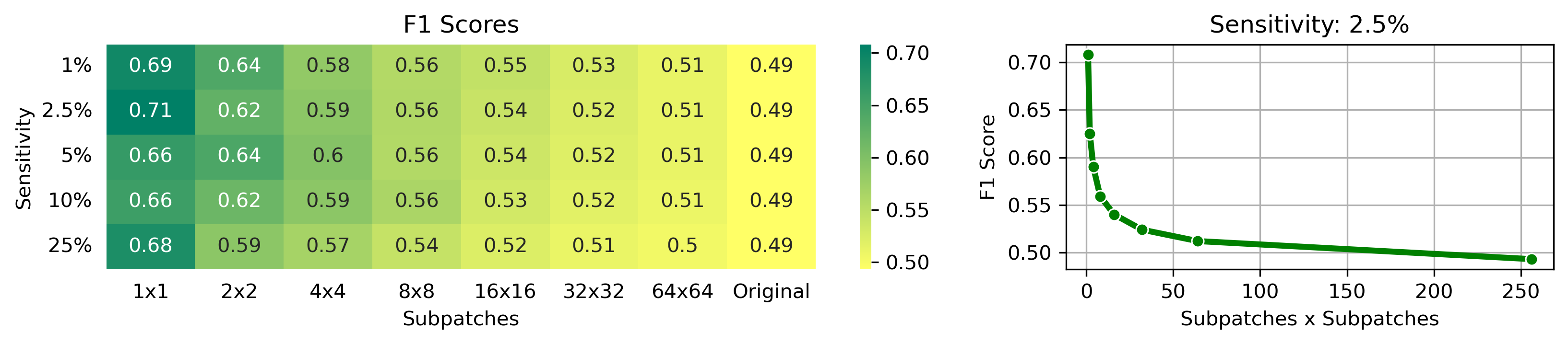}
	\caption{F1 scores for the flood prediction model under a variety of combinations of ARLA sensitivity and subpatches parameters.}
	\label{fig7}
\end{figure}

\begin{figure}[!h]
    \centering
	\includegraphics[width=0.90\textwidth]{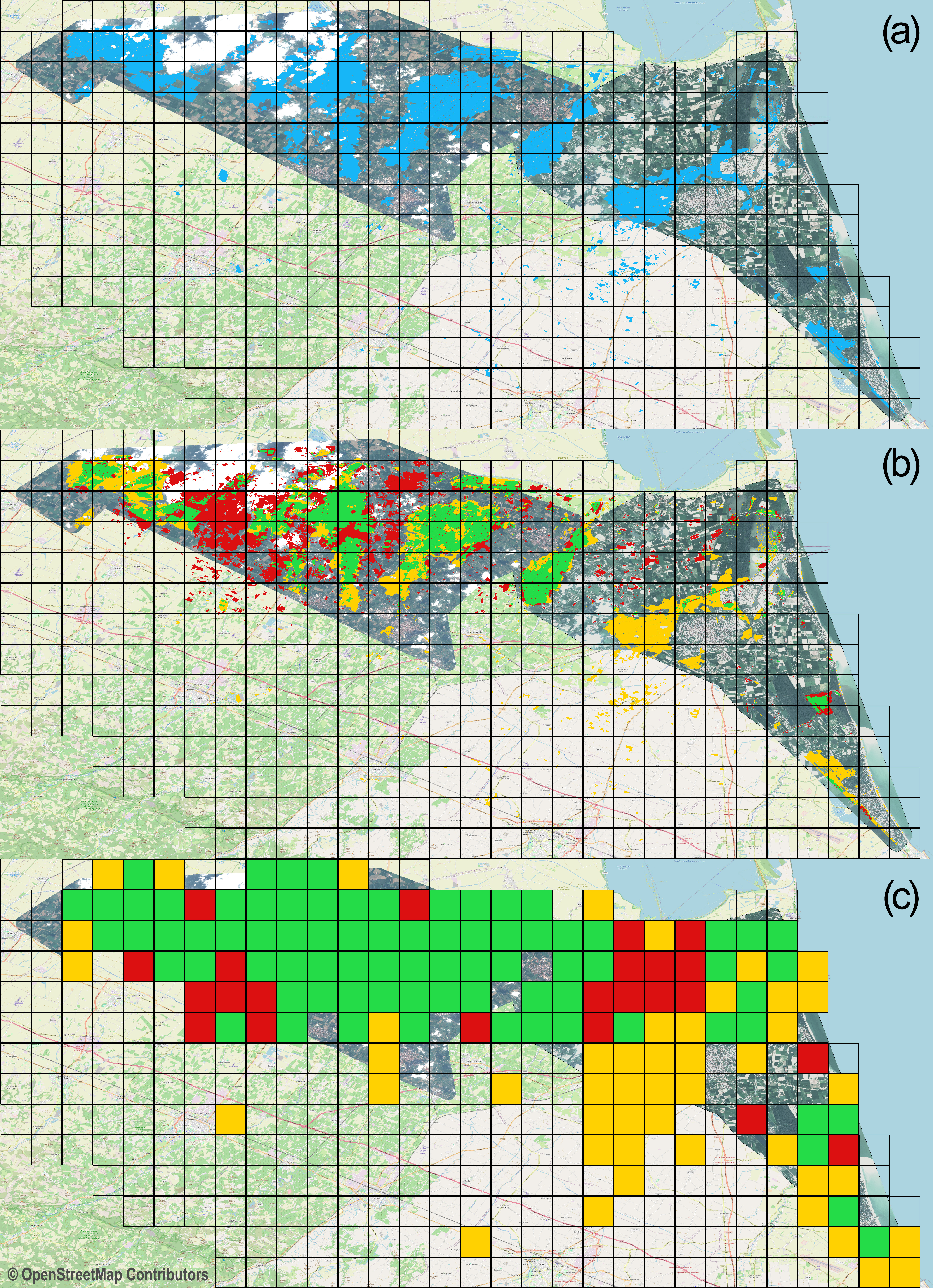}
	\caption{Unreliable labels of a real flood event (a). The performance of a flood prediction model is compared before (b) and after (c) ARLA has been applied. Green shows where the prediction matches the label, red over-prediction, and yellow under-prediction.}
	\label{fig6}
\end{figure}

Overall, even at this coarsest possible resolution ($1 \times 1$ subpatches), the model behaviour is made more comprehensible, as it is now apparent that the model tends more to under-predict than over-predict floods, whereas the inverse was previously concluded from evaluating on the original segmentation. However the noisy automated labels generated in the bottom-right of the image have not been entirely eliminated after applying ARLA, thus a higher sensitivity threshold could be used if the intention was to focus on only strong signals from unambiguously flooded regions. Model performance can hence be reported over a range of sensitivities and number of subpatches, as in figure \ref{fig7}, which uses the F1 score as the example metric. The graph in figure \ref{fig7} also shows how the model performance over the resolution (at a fixed sensitivity) follows a power law distribution, varying between 0.5 and 0.7 F1.

\newpage
\section{Discussion}\label{discussion}
In the results, we have seen how ARLA is able to handle unreliable label errors, such as cloud cover, without obscuring either the model's strengths or weaknesses, but rather illuminating the model's behaviour. Furthermore, because ARLA is applied at inference time, not during training, the model does not need to be re-trained for every variation, improving accessibility. As described in section \ref{related_work}, deep learning models are already relatively resilient to training on unreliable labels, (and many methods exist to improve their robustness further), as the model can learn to extract the patterns from the noise. However noisy labels inherently and unavoidably distort model evaluation, thus motivating the development of approaches such as ARLA.

Generating and analysing performance metrics tables and graphs of parameter combinations, such as in figure \ref{fig7}, can assist in selecting the most appropriate parameters to report results by. ARLA was intentionally designed to have few adjustable parameters, in order to deter overfitting the labels to the model's prediction. They are not intended as hyper-parameters to be optimised to maximise model performance, but rather can be varied to evaluate the performance of the model under different conditions and at different resolutions. For example, the model in figure \ref{fig6} achieves an F1 score of 0.49 at the original resolution of 10m per pixel, and an F1 score of 0.71 at a resolution of 2.56km/patch ($1 \times 1$ subpatches and sensitivity of 2.5\%).

The resolution parameter can also be adapted to the size of the label error, the precision of the labels, or the level of the label noise. In figure \ref{fig4}b, $20 \times 20$ subpatches were used, as this resolution best fit to the size of the cloud cover holes in the label. Given the range of a label error size, any model variance that is also within this range thus cannot be confidently penalised, however larger model deviances persist even after ARLA has been applied, as demonstrated by figure \ref{fig4}a. Any smaller model errors may be concealed via aggregation, but if they fall within the expected label error variance, then these are not guaranteed to be model errors anyway. Furthermore, if labels are not precise, then reporting model performance at the highest original resolution is misleading, because it implies a precision at that resolution which may not exist. Nevertheless, estimating the size of the label error may be difficult without access to the original annotators' knowledge or reference data. In this case, results can simply be transparently reported at multiple resolutions, as in figure \ref{fig7}.

The sensitivity parameter can be tuned to the level of label noise and how meaningful each label is. A low sensitivity helps to fill in the gaps of inconsistent labels, prioritising avoiding false negatives rather than false positives. With very noisy labels however, the sensitivity can be increased so that only the strongest signals are retained, for example, the application demonstrated in figure \ref{fig6}c may have benefitted from a higher sensitivity to exclude all of the noisy auto-generated label patches.

The ARLA method is especially useful for evaluating flood prediction segmentation models, because their primary aim is to firstly forecast whether a flood will occur at all in a region, and then additionally which areas are likely at risk. This task can therefore be achieved even with a much coarser label aggregation. The aggregated metrics also may be more reliable if the flood label is annotated inconsistently with ambiguous borders, thus attempting to segment and evaluate these at a high resolution is not particularly meaningful anyway.

On the application examples, ARLA performed better than the two baseline methods, particularly with regards to handling large label errors, true model errors, and correct areas in the labels. However, these comparisons will be extended to evaluate the performance and the behaviour of ARLA on a broader benchmark dataset. Future works on ARLA should also explore further applications in addition to flood prediction models, to better investigate the benefits and limitations of the method. This could include a systematic analysis of the effect of different subpatches and sensitivity parameter values, as well as the development of a standardised method to select the ARLA resolution based on the label error size. ARLA is a powerful new approach for evaluating segmentation models on even highly unreliable labels, but future works will be necessary to learn how to most effectively leverage its capabilities.

\section{Conclusion}\label{conclusion}
Labels are critical for both training and evaluating machine learning and deep learning models, however they often contain errors, particularly in segmentation datasets. Labels may be unreliable, inconsistent, noisy, or ambiguous at class borders, and these errors may be impossible to clean, even with the resources or knowledge of the original annotators. Although many approaches to train models on weak labels currently exist, no method has been previously developed to evaluate segmentation models on unreliable labels. 

To fill this gap we therefore developed the `Adaptive Resolution Label Aggregation' method, or `ARLA'. ARLA adapts the resolution of both the label and the model prediction, in order to minimise mislabelling errors and extract a clearer signal of both the information encapsulated by the label and of the model performance. In an application to flood prediction, ARLA was able to overcome issues with inconsistent labelling of forested areas, and errors in labels within regions of heavy cloud cover, better than baseline comparison methods. In a further example, evaluating a flood prediction model on highly error-prone labels obscured the true model performance. By applying evaluation metrics to compare the ARLA aggregated label and model prediction instead, even at a very coarse resolution, ARLA was able to better analyse the model behaviour, revealing how the model struggled with recall, rather than precision, as was suggested by the original segmentation metrics.

The ARLA parameters can be varied in order to evaluate a model at different combinations of sensitivities and numbers of subpatches (resolutions). The parameters can also be adapted to the size of the label error, the precision of the labels, or the level of the label noise. With this method, model performance can be reported fairly by minimising the label error without obscuring the true model error, thus achieving a reliable approach to evaluate models on unreliable labels.

\section*{Data Sources and Licensing}\label{data}
Copernicus Sentinel-2 data retrieved from the CDSE Sentinel Hub.  \url{https://dataspace.copernicus.eu/analyse/apis/sentinel-hub}.
\\ \\
Copernicus Emergency Management Service On-Demand Mapping \url{https://mapping.emergency.copernicus.eu/} from the Directorate Space, Security and Migration, European Commission Joint Research Centre (© European Union, 2012-2026).
\\ \\
The OpenStreetMap (OSM) data is distributed under the Open Database License (ODbL). \url{https://www.openstreetmap.org/copyright}.
\\ \\
The Cityscapes dataset is available at \url{https://www.cityscapes-dataset.com/}.
\\ \\
All code and reproducible figures, tables, and examples are available in the following GitHub repository: \url{https://github.com/Natasha-R/ARLA-Evaluating-on-Unreliable-Labels}

\bibliographystyle{plainnat}
\bibliography{references}
\end{document}